%% file: main.tex
\documentclass[10pt,twocolumn,letterpaper]{article}

\usepackage{3dv}
\usepackage{times}
\usepackage{epsfig}
\usepackage{graphicx}
\usepackage{amsmath}
\usepackage{amssymb}
\usepackage[font=small]{caption}
\usepackage{subcaption}

\usepackage{graphics} 
\usepackage{mathptmx} 
\usepackage{multirow}
\usepackage{epstopdf}
\usepackage{booktabs}
\usepackage{makecell}

\newcommand{\hide}[1]{}
\renewcommand{\vec}[1]{\mathbf{#1}}
\newcommand{\mat}[1]{\mathbf{#1}}

\usepackage[pagebackref=true,breaklinks=true,letterpaper=true,colorlinks,bookmarks=false]{hyperref}

\threedvfinalcopy 

\def\threedvPaperID{96} 

\ifthreedvfinal\fi
\begin{document}

\title{3D Object Discovery and Modeling Using Single RGB-D Images Containing Multiple Object Instances}
\author{Wim Abbeloos$^1$, Esra Ataer-Cansizoglu$^2$\thanks{Corresponding authors: {\tt \{cansizoglu,taguchi\}@merl.com}}, Sergio Caccamo$^3$, Yuichi Taguchi$^{2*}$, Yukiyasu Domae$^4$ \\ \\
$^1$KU Leuven, Belgium\\
$^2$Mitsubishi Electric Research Labs, Cambridge, MA, USA   \\
$^3$KTH Royal Institute of Technology, Stockholm, Sweden   \\
$^4$Mitsubishi Electric Corporation, Japan
}
\maketitle

\thispagestyle{empty}
\pagestyle{empty}

\input{abstract}

\input{intro}

\input{relatedwork}

\input{method}

\input{experiments}

\input{discussion}

\section*{Acknowledgments}
This work was done at and supported by Mitsubishi Electric Research Laboratories. We thank the anonymous reviewers for their helpful comments.  W.A. thanks EAVISE, KU Leuven for the travel support.

\addtolength{\textheight}{-5.5cm}

{\small
\bibliographystyle{ieee}
\bibliography{SLAM,MOI3D}
}


\end{document}

%% file: abstract.tex
\begin{abstract}

Unsupervised object modeling is important in robotics, especially for handling a large set of objects. We present a method for unsupervised 3D object discovery, reconstruction, and localization that exploits multiple instances of an identical object contained in a single RGB-D image. The proposed method does not rely on segmentation, scene knowledge, or user input, and thus is easily scalable. Our method aims to find recurrent patterns in a single RGB-D image by utilizing appearance and geometry of the salient regions. We extract keypoints and match them in pairs based on their descriptors. We then generate triplets of the keypoints matching with each other using several geometric criteria to minimize false matches. The relative poses of the matched triplets are computed and clustered to discover sets of triplet pairs with similar relative poses. Triplets belonging to the same set are likely to belong to the same object and are used to construct an initial object model. Detection of remaining instances with the initial object model using RANSAC allows to further expand and refine the model. The automatically generated object models are both compact and descriptive. We show quantitative and qualitative results on RGB-D images with various objects including some from the Amazon Picking Challenge. We also demonstrate the use of our method in an object picking scenario with a robotic arm.

\end{abstract}

%% file: intro.tex
\section{Introduction}

\begin{figure}
\centering
\includegraphics[width=0.45\textwidth]{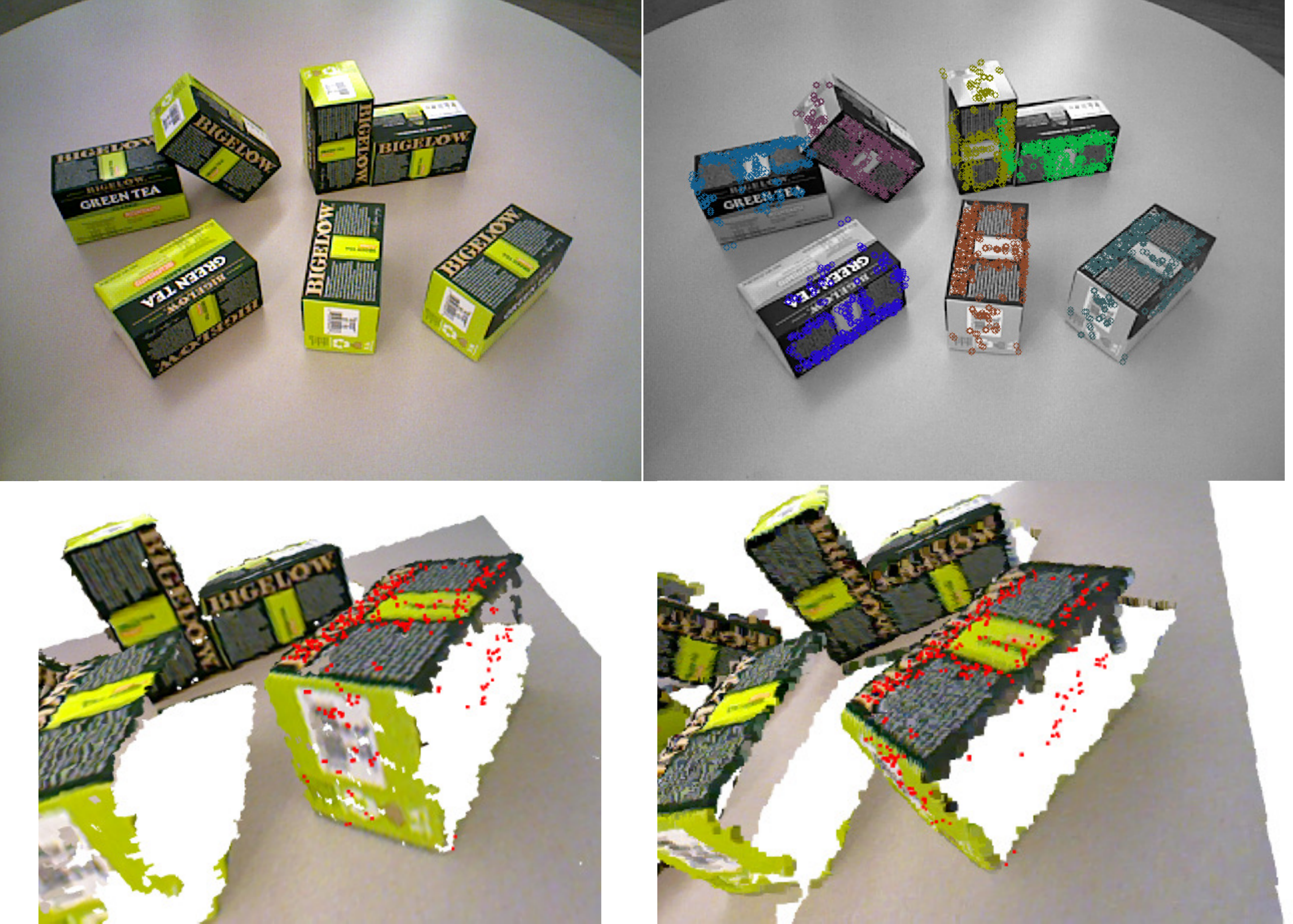}
\caption{Given a single RGB-D image containing multiple instances of the same object (top-left), our method automatically discovers the object and localizes the multiple instances by grouping a set of features (top-right). A 3D model of the object is also recovered by registering the features from the multiple instances into a single coordinate system (bottom). The registered features are denoted as red dots, overlaid on the colored 3D point cloud of the scene. Note that some of the features appear on the missing face of this specific object instance, indicating that they are recovered from some other object instances.}
\label{fig:teaser}
\end{figure}

Object model generation is crucial for robotic manipulation. Typical object detection and localization methods have a separate supervised stage where they learn and build object models. However, the types of objects a robot needs to interact with can expand and change rapidly, such as new items arriving at a warehouse as seen in the scope of Amazon Picking Challenge~\cite{apc}. On the other hand, in many situations, objects appear in multiple copies. This paper exploits this fact and presents a method for discovering and modeling an object from a single RGB-D frame in which the object appears in multiple copies. The recurrent patterns found in the single frame can be used to automatically discover the object, and the various viewpoints of different instances can provide valuable information for object model generation.

Our only assumption is the existence of at least two instances of an object in the single RGB-D image.
We do not use any prior knowledge about the number, shape, and appearance of the object. Thus, the object can appear in a cluttered scene or the image can contain multiple instances of different objects.
Our method performs on-the-fly object model generation, while detecting and localizing the instances of the reconstructed object in the given image. Thus, it enables online robot manipulation using only a single-shot image.

Our technique employs a sparse feature representation, as shown in Figure~\ref{fig:teaser}. Therefore, the problem can be seen as finding groups of features that correspond to different instances of the object. To solve this grouping problem we make use of the following information:
\begin{enumerate}
\item \textit{Appearance similarity:} Pairs of features that come from the same location of two instances should be similar.
\item \textit{Geometric similarity:} Two groups of features corresponding to each other based on appearance similarity should have the same in-group geometric distribution. In other words, there exists a single transformation that would transfer and align the positions of features in one group to the positions of corresponding features in the other group.
\end{enumerate}
We employ the appearance and geometric constraints jointly. Furthermore, we avoid the use of depth segmentation and spatial closeness to decide whether features are coming from the same instance, as the objects might be touching with each other or occluding one another.

We look for recurrent patterns in the image using both geometric and appearance similarity following the sparse feature representation. First, we extract keypoints and match them based on the descriptor similarity. We then find triplets of keypoints matching with each other using several geometric criteria, which are defined for pairs and triplets of the matched keypoints and are invariant to the 6-degree-of-freedom (6-DOF) transformations. Each of the matched triplets provides a 6-DOF transformation, which is a candidate of the relative pose between two instances of the object, but might be an outlier. Thus, in the second stage we cluster the relative poses associated with each triplet match and find clusters supported by many triplets. The matches that appear in the same cluster are likely to belong to the same pair of objects. Thus, in the third stage we generate an initial model based on the clustering results. Lastly, the generated model is used in a RANSAC framework in order to detect additional instances among the remaining keypoints, which can yield further expansion and enrichment of the generated model.  Once a model has been discovered, it can be used to detect the corresponding object even if only one instance is present.


\subsection{Contributions}

The main contributions of this paper are as follows:
\begin{itemize}
\item We present a method for unsupervised object discovery and modeling from a single RGB-D image containing multiple instances of the same object.
\item We propose an efficient grouping algorithm that generates a set of relative pose candidates using triplets of keypoint matches and then clusters them to find each instance of the object and their relative poses.
\item We show experimental results using several objects and demonstrate an application of our method to object picking.
\end{itemize}

%% file: relatedwork.tex
\subsection{Related Work}

Object discovery has been investigated using a variety of approaches.  Some are based on geometric and/or color segmentation~\cite{Papon13CVPR}\cite{firman2013learning}, which typically rely on strong assumptions of the scene or the objects (e.g., the objects are placed on a table) and do not exploit multiple instances. Another segmentation-based approach based on shape analysis using compactness, symmetry, smoothness,  and  local  and  global  convexity of segments and their recurrence is proposed in \cite{karpathy2013object}.  Since these methods suffer from over and under segmentation, especially in scenes with a lot of clutter, they are not suitable solutions to our problem.  Other methods gather information over time (thus they are not single-shot approaches)~\cite{dharmasiri2016mo}\cite{Ma2014} and some assume that objects will be displaced or removed~\cite{herbst2011rgb}\cite{mason2012object}\cite{Caccamo2017}.

A closely related field is the detection of repetitive patterns~\cite{leung1996detecting}\cite{pauly2008discovering}\cite{wu2010detecting} in images.  These methods, however, depend on the organized appearance of the structure elements, while in our case, object instances may appear in random poses.  Other related problems are co-segmentation~\cite{chu2010momi} and unsupervised detection of object categories~\cite{tuytelaars2010unsupervised}.

Unsupervised detection of multiple instances of objects in RGB images has been studied in~\cite{ferrari2006simultaneous}\cite{cho2009feature}\cite{cho2010unsupervised}\cite{liu2013grasp}.  These methods also use geometric and appearance information, which is then used for clustering or combinatorial optimization.  They operate either on matched points, or matched pairs of points, while our method uses matched triplets.  Moreover, they use only RGB information and not depth, and do not reconstruct a 3D model or estimate 6-DOF pose.  We compare our method with~\cite{cho2009feature} in experiments.

Some object detection methods for robotics applications have been proposed that take into account multiple object instances.  A sparse 3D object model created by using structure from motion, which requires multiple frames, is used in \cite{collet2009object}.  This enables the detection and pose estimation of multiple object instances in an RGB image.  Similarly, in \cite{grundmann2010robust}, a sparse 3D model is first manually created from multiple images, after which they detect the model using a stereo camera system. Note that these systems require one to build a model first, while ours does not.

%% file: method.tex
\section{Method}

The goal of this work is to discover, model, and localize an object in a scene without any prior knowledge. The input is a single RGB-D frame, including a color (or grayscale) image and a depth map of the scene. We use sparse 3D feature points throughout our pipeline, and thus ignore pixels that have invalid depth measurements.

Our method consists of four main steps. In the first step, we extract keypoints and generate triplet matches based on the descriptor similarity and several geometric criteria that are invariant to the 6-DOF transformations. Second, we cluster triplet matches based on their relative poses as we expect to see geometric similarity among groups of features. Third, we generate an initial model using the clustering results. At the fourth step, the initial model is used to detect additional object instances in the remaining set of features that have been considered outliers in the clustering step, which can further enhance the object model. Each of the four steps is detailed in the following subsections.

\hide{The goal in this work is to discover, model, detect and estimate the pose of 3D objects in a scene without any prior knowledge. The input consists of an RGB (or gray scale) image and a 3D point cloud of the scene. It is assumed that multiple instances of an object are visible.
Correspondences among these object instances are used to automatically build a 3D object model, which can be used
to detect other instances of the same object at a later stage.}

\subsection{Matching Triplets of Keypoints}

In the first step, our goal is to generate triplets of keypoint matches, each of which provides a candidate of the relative pose between two instances of the object.
We use SIFT~\cite{Lowe04IJCV} to detect and describe keypoints. This results in a set of $N$ keypoints that have valid depth measurements. Every keypoint in this set is compared to all others to find its most similar keypoint. The similarity measure used is the Euclidean distance between the $128$ dimensional feature descriptors. We also threshold the Euclidean distance such that we maintain $M \leq N$ keypoint matches for the following processes.

\begin{figure}
\centering
\includegraphics[width=0.3\textwidth]{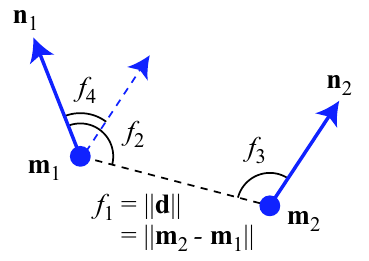}
\caption{Two surface points $\mathbf{m}_{i}$ and their normals $\mathbf{n}_{i}$ determine a point pair feature.}
\label{fig:ppf}
\end{figure}

Based on appearance similarity, we expect that two instances of an object have similar keypoints. However, the single keypoint matches are not robust enough, include many outliers, and do not provide the relative pose between the two instances. Thus triplets of keypoint matches are used to be robust to outliers and to obtain the relative pose using three 3D point registration~\cite{Umeyama91PAMI}.
For each combination of three keypoint matches, we need to consider cases where matches are reversed except for symmetric cases (we thus keep 4 out of 8 candidates). This results in 
a total of $4\begin{pmatrix}
M\\
3
\end{pmatrix} = \frac{2M (M-1) (M-2)}{3}$ possible triplets. We try to select correct triplets based on the following geometric criteria invariant to the 6-DOF transformations:
%
%
%
\begin{itemize}
	\item{Point pair feature similarity: Point pair features \cite{Drost10CVPR}\cite{Choi12Voting}\cite{Abbeloos16ppf} describe the relative position and orientation of points on the surface of an object (Figure~\ref{fig:ppf}).  For two points $\mathbf{m}_{1}$ and $\mathbf{m}_{2}$ with normals $\mathbf{n}_{1}$ and $\mathbf{n}_{2}$, with $\mathbf{d} =  \mathbf{m}_{2} - \mathbf{m}_{1}$ the feature  $\mathbf{F}$ is
\begin{equation}
\mathbf{F}(\mathbf{m}_{1} , \mathbf{m}_{2} ) = ( \lVert \mathbf{d} \rVert , \angle(\mathbf{n}_{1} , \mathbf{d}), \angle(\mathbf{n}_{2} , \mathbf{d}), \angle(\mathbf{n}_{1} , \mathbf{n}_{2} )) ,
\end{equation}
where $\angle(\mathbf{a}, \mathbf{b}) \in [0 \quad \pi]$ denotes the angle between two vectors. Let $\vec{l}_1$ and $\vec{l}_2$ be keypoints matching with $\vec{m}_1$ and $\vec{m}_2$ respectively. We compute the difference of point pair features between the matches as $\vec{F}(\vec{m}_1, \vec{m}_2)-\vec{F}(\vec{l}_1, \vec{l}_2)$ and apply distance and angle thresholds to the calculated difference to filter out incorrect correspondences.}
\item{Sidedness: We check whether the third point of the triplet falls on the same side of the line defined by the other two points to avoid reflections~\cite{ferrari2006simultaneous}. For the triplet $P$, let us denote the cross product of the edges $\vec{d}_i$ and $\vec{d}_j$ as $\vec{v}_{i,j}(P) = \vec{d}_i \times \vec{d}_j$. We discard a triplet match if any of the two corresponding $\vec{v}$ vectors are in opposite directions, i.e., discard the triplet match between $P$ and $Q$, if $\exists i,j$ such that $\left \| \frac{\vec{v}_{i,j}(P)}{\left \| \vec{v}_{i,j}(P) \right \|} + \frac{\vec{v}_{i,j}(Q)}{\left \| \vec{v}_{i,j}(Q) \right \| }\right \|  < \epsilon $.
}
\item{Minimum triangle edge length and maximum acuteness: To ensure the found corresponding triangles will yield sufficiently accurate transformation estimations, triangles generated with closely located keypoints are removed.  This is done using a minimum triangle edge length and maximum angle acuteness threshold.}
\item{Overlapping triangles: We omit the triplet match if the two triangles are overlapping, as they would more likely be coming from the same instance.
}
\end{itemize}
Since the point pair feature similarity can be computed for pairs of keypoint matches, we first use this criterion for efficient pruning of incorrect pairs and then use the other criteria for selecting correct triplets. Also, in order to get a balanced distribution of keypoints among triplet matches, a keypoint can appear at most $L$ times in the generated triplet matches. The thresholds and parameters used in this study are given in Section~\ref{sec:exp}.


\subsection{Clustering}

For each of the triplets obtained in the first step, a 6-DOF pose that transforms the triangle to its corresponding triangle is estimated. Let $P=(\vec{p}_1, \vec{p}_2, \vec{p}_3)$ and $Q=(\vec{q}_1, \vec{q}_2, \vec{q}_3)$  denote two matching triangles where $\vec{p}_i,\vec{q}_i\in \mathbb{R}^3$ are 3D positions of the keypoints. The calculation of the pose results in the transformation $\mat{T_{p,q}}\in SE(3) $ that consists of a rotation matrix $\mathbf{R}\in SO(3)$ and a translation vector $\mathbf{t}\in \mathbb{R}^3$ such that $\vec{q}_i = \mat{T_{p,q}}(\vec{p}_i)=\mat{R}\vec{p}_i + \mathbf{t}$.  These transformations are clustered using the DBScan~\cite{ester1996density} algorithm to discover sets of triplets with similar transformations. DBScan is a density based clustering method, which only requires a single input parameter for the maximum distance between two instances that are allowed to be clustered together\hide{ (and optionally the minimal number of items that need to be clustered together to consider it as a valid cluster)}. During clustering, we exploit sum of 3D point-to-point distances as the distance between two triplets. For symmetry, the distance is computed bidirectionally. Thus, the distance between two matching triplets $(P, Q)$ and $(A, B)$  based on the respective transformations $\mat{T_{p,q}}$ and $\mat{T_{a,b}}$ is
\begin{equation}
D( (P,Q), (A,B) ) = \sum_{i}\| \mat{T_{p,q}}(\vec{a}_i)-\vec{b}_i \| + \sum_{i}\| \mat{T_{a,b}}(\vec{p}_i)-\vec{q}_i \| .
\label{eq:distance}
\end{equation}
%
%

The output of clustering can contain the same pair of instances in two different clusters with associated poses as inverse of each other. Hence, if such clusters are found, one of them is inverted and the clusters are merged. The transformation for each cluster is then recalculated considering all sets of corresponding triplets in the cluster.

\subsection{Initial Model Creation}

\begin{figure}
\centering
\includegraphics[width=0.45\textwidth]{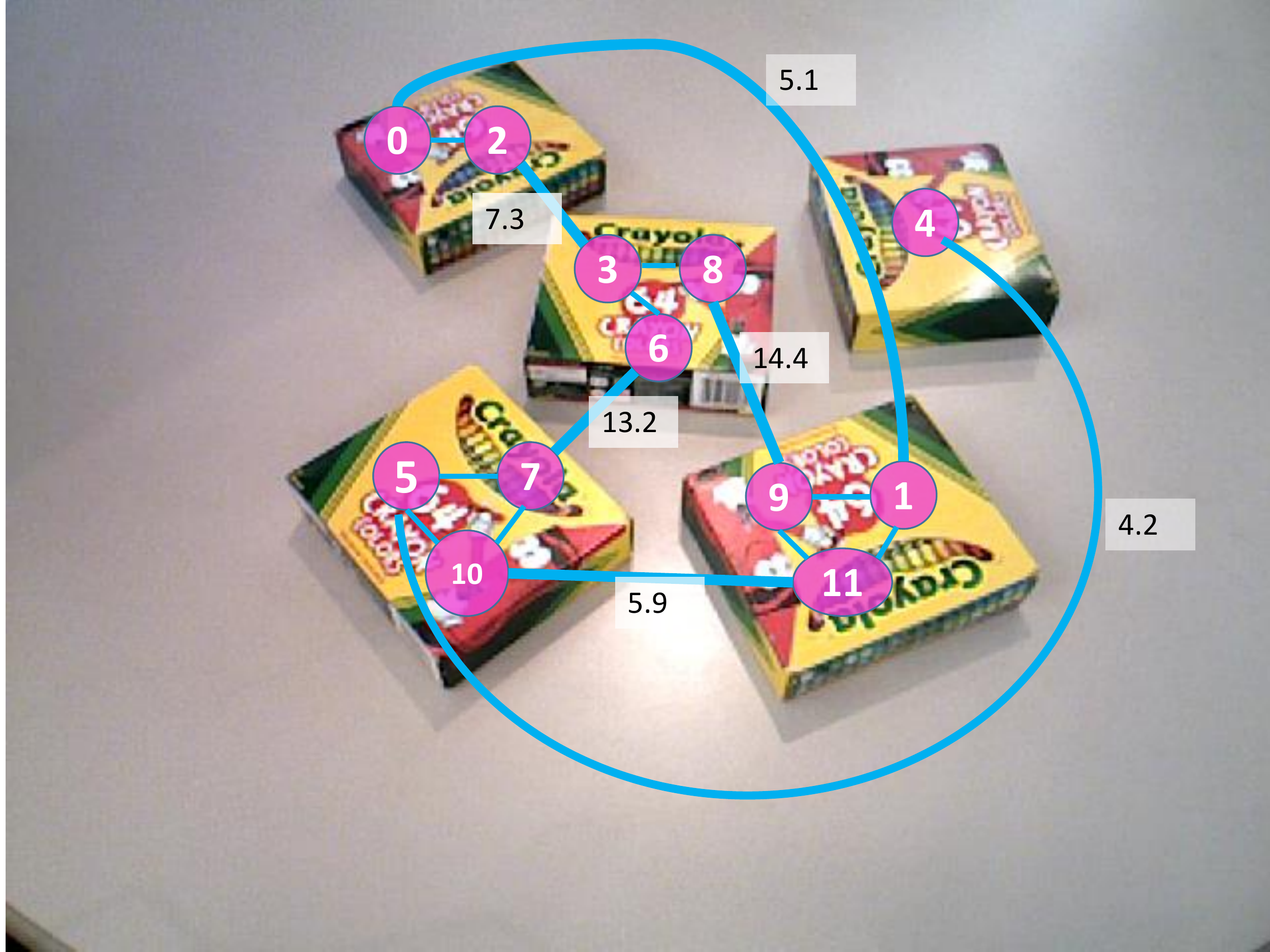}
\caption{An overlay shows the graph representing clustered sets of points (vertices) and their relations (edges).  Note there are two types of relations between sets of points: sets that are clustered together because they contain similar triplets, and sets that are connected because they have points in common.  The edges representing matched triangles have a label showing the distance (the distance error obtained with the transformation between them).}
\label{fig:graph}
\end{figure}

The clustering procedure results in sets of points that belong to the same object instance and are matched to another object instance\hide{ (within a cluster, they are all matched to the same object)}. In other words, each cluster can be seen as two sets of points, where one set can be aligned with the other set using the transformation of the cluster. Some of these sets may have keypoints in common with other sets. Thus, the clustering result can be represented as a graph where nodes correspond to sets of points and edges correspond to the distance between sets based on the transformation of the cluster associating the two sets (Figure~\ref{fig:graph}). If two sets have points in common, then the transformation between them is identity and the connecting edge is set to have a small preset weight $\delta$.

The resulting graph can have multiple connected components, since the scene can contain multiple instances of various types of objects. In order to create a model for each connected component, we first decide which node will be the reference frame where all sets will be transformed to. We pick the node representing the set of points with the largest number of matches and common points as the reference. All other sets of points that are connected to it are transformed to the reference frame by applying a series of transformations.  The optimal series of transformations for every set is found by searching for the shortest path to the reference frame using Dijkstra's algorithm~\cite{dijkstra1959note}.

The 3D object model consists of all points transformed to this common reference frame, and associated with their original keypoint descriptors. This process generates an object model for each connected component in the graph, hence it might yield multiple models, each containing points from all sets connected to their reference set. For each generated model, we apply a final bundle adjustment to refine landmark locations and instance poses.

\subsection{Additional Instance Detection}

After creating a set of object models, every model is compared to all others to verify whether they truly are distinct objects, or whether their correspondence was simply missed by the earlier steps (this is possible because we enforce a unique match between keypoints in our first step, instead of considering all possible matches).  For each model, we perform detection between the model and the sets of points from the other connected components of the graph\hide{set of points that do not contribute to the model}. This is performed by a correspondence search via descriptor similarity and a geometric verification by a 3-point RANSAC registration.

We also try to detect any remaining instances that had not been matched before.  We use the remaining keypoints that are not associated with any of the nodes in the graph to avoid matching the model to the previously detected instances.
In the RANSAC registration, we sample three scene points so that they are within the diameter of the model.

In both cases, the RANSAC estimates an initial transformation using the three points and counts the number of inliers (the percentage of matched points that, when transformed, are within a certain distance of their corresponding points).  RANSAC succeeds if the inlier ratio is larger than a certain threshold. The transformation is then re-estimated based on the inliers of the most successful attempt. In the case of a successful RANSAC, the model is merged with the other model or the points selected as inliers from the remaining keypoints.

%% file: experiments.tex
\section{Experiments and Results}
\label{sec:exp}

An ASUS Xtion Pro Live RGB-D camera was used to acquire a dataset of $234$ VGA ($640 \times 480$) resolution color and depth images.  The depth image was converted to a 3D point cloud and transformed to the RGB camera's reference frame.  This means every valid measurement point has both a 3D coordinate and a color value.

We classified the captured scenes into four different scenarios: scenes containing either a single or multiple object types (denoted by $S$ and $M$), and scenes with or without clutter and occlusion (denoted by easy $_{ez}$ and difficult $_{d}$).  Each scene contained two to nine instances per object type. We used objects with various shapes and sizes and varying amounts of texture.
The number of images per scenario is given in Table~\ref{table:dataset_numbers}.
Qualitative and quantitative results are given for the different scenarios.  The dataset is available at~\url{ftp://ftp.merl.com/pub/cansizoglu/ObjectDiscovery3DV2017.zip}, and includes object annotations.

\begin{table}[t]
\centering
\begin{tabular}{cccc}
 & \# img (pair)& \# img ($>2$)& \# img (total) \\
$S_{ez}$ & 54& 29& 83\\
$S_{d}$ & 57& 0& 57\\
$M_{ez}$ & 49& 3& 52\\
$M_{d}$ & 30& 12& 42\\
\hline
 & 190& 44& 234\\
\end{tabular}
\caption{The number of images containing only pairs of instances, the number of images containing more than two instances, and the total number of images, per scenario.}
\label{table:dataset_numbers}
\end{table}

Our method is compared to an RGB object discovery algorithm~\cite{cho2009feature} that uses feature matching with a novel pairwise dissimilarity measure and hierarchical agglomerative clustering to find pairwise matches between sets of points.  The dissimilarity measure they propose consists of a photometric and a geometric term.  The photometric term is simply the Euclidean distance between the points' SIFT descriptor vectors, which is also used in our method.  The geometric term is used to determine a pairwise dissimilarity between two corresponding pairs.  It uses the homography of the first correspondence to transform the points of another and vice versa.  The final geometric term is the average of the transformation errors.  The total pairwise dissimilarity is a linear combination of both terms.  We used the source code available on the authors' website with the default parameters, as changing them did not improve the results.

The following parameters were used in these experiments to eliminate incorrectly matched triplets: a $5mm$ and $35$ degrees threshold for the point pair feature difference.  Each edge of the triangle should be at least $10mm$ and at most $125mm$ and each angle should exceed $10$ degrees.  Maximum value of the distance between two samples in clustering was set to $35mm$, while we discarded clusters with less than $14$ samples.  In detection, we used a RANSAC inlier threshold of $5mm$. A RANSAC was recalled as successful when there were at least $5$ inliers and the inlier ratio was more than $12.5\%$.  The average running time was 809ms with a C++ implementation on a standard PC.

\begin{figure}[t]
\centering
\includegraphics[width=0.35\textwidth]{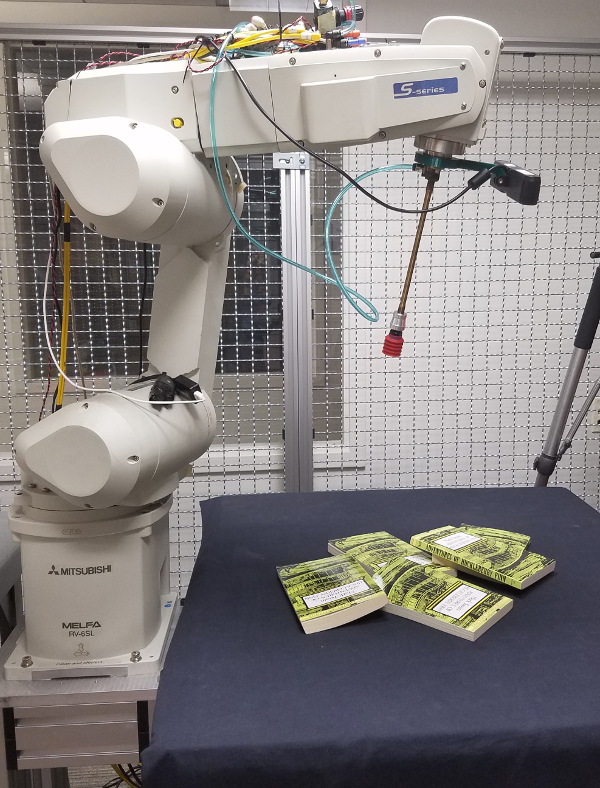}
\caption{Robot arm used for object picking.}
\label{fig:robot}
\end{figure}

We also demonstrate the use of our algorithm in an object picking scenario with a robotic arm, where multiple instances of the same object are visible (Please see supplementary video). We mounted an ASUS Xtion sensor on the robot arm and picked up objects using a vacuum gripper as seen in Figure~\ref{fig:robot}.

\begin{figure*}[tp]
\centering
\includegraphics[width=1\textwidth]{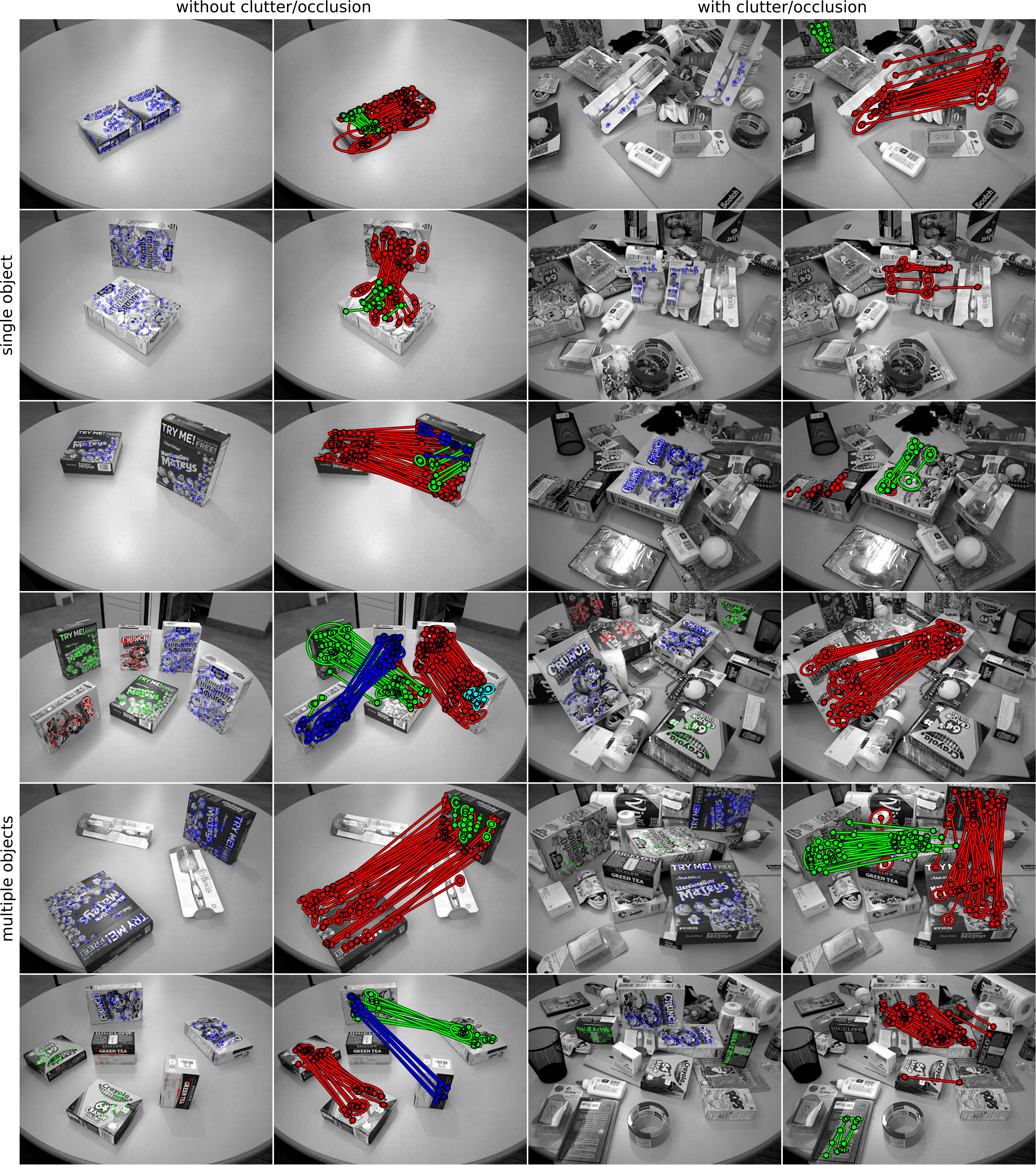}
\caption{Some of the scenes from the proposed dataset from the four scenarios: single/multiple objects (top/bottom) and with/without clutter and occlusion (left/right).  Overlaid is a visualization of the results for our method (first and third columns) and the results for~\cite{cho2009feature} (second and fourth columns).  The quantitative results are summarized in Table~\ref{table:results}.}
\vspace{1cm}
\label{fig:dataset}
\end{figure*}

\begin{figure*}[t]
\centering
\begin{subfigure}[t]{.3\textwidth}
  \centering
  \includegraphics[width=.98\linewidth]{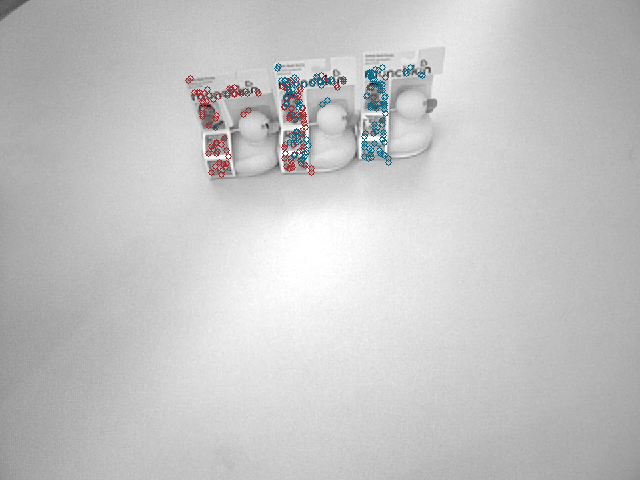}
  \caption{Because our algorithm searches for the largest recurrent pattern, object instances occurring in an organized way are merged into one object, resulting in object models containing two instances.  Each object instance is shown in a random color.}
  \label{fig:duckthree}
\end{subfigure}%
\quad
\begin{subfigure}[t]{.3\textwidth}
  \centering
  \includegraphics[width=.98\linewidth]{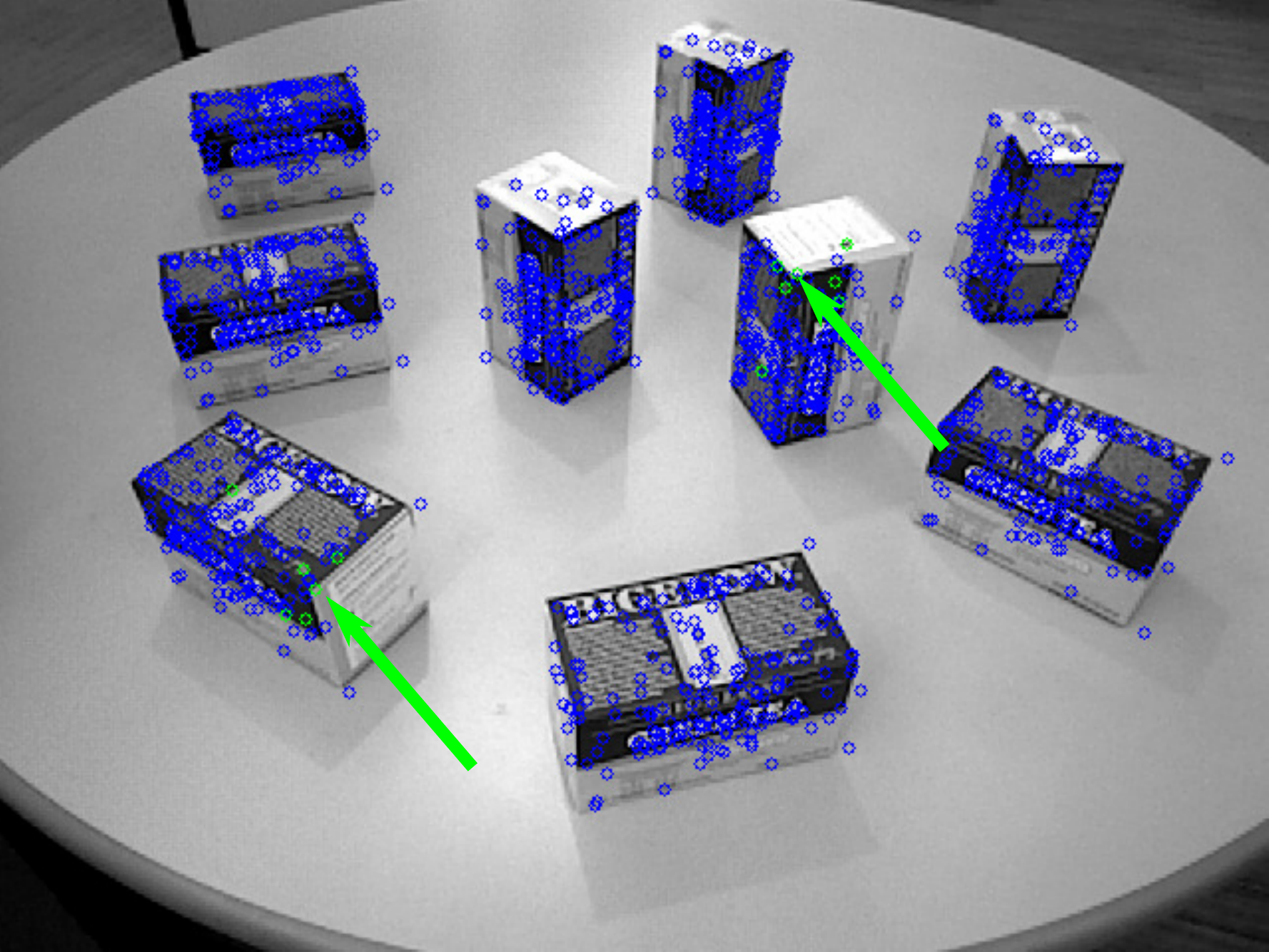}
  \caption{While a correct object model was created for the object, and all instances were found correctly, a separate partial model was discovered in two instances (denoted by green arrows pointing to green model points).  The second model is considered a false positive, since these points should have been a part of the first model.}
  \label{fig:falsepos}
\end{subfigure}%
\quad
\begin{subfigure}[t]{.3\textwidth}
  \centering
  \includegraphics[width=.98\linewidth]{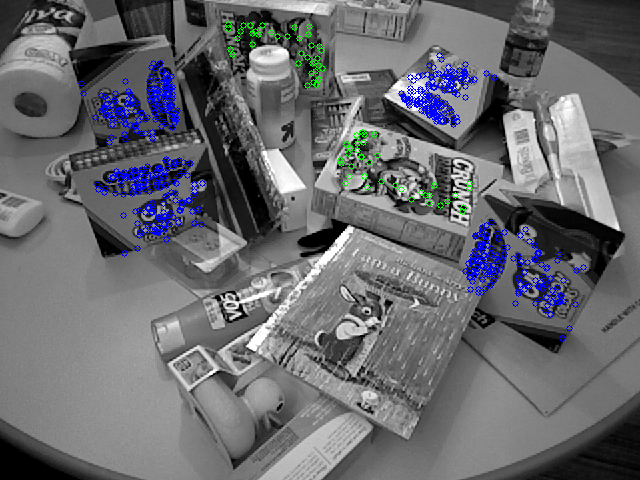}
  \caption{The object model in green was successfully discovered, and its instances located, despite the high degree of occlusion by other objects.}
  \label{fig:sub3}
\end{subfigure}
\caption{A few remarkable results with additional comments.}
\label{fig:test}
\end{figure*}

\subsection{Qualitative Results}

Some results on the proposed dataset on the four different scenarios are shown in Figure~\ref{fig:dataset}.    For our method the image is overlaid with the transformed object model (with one color per object type).  For the comparison method, clusters of matching features are shown (each cluster having a different color).  This figure contains only scenes with two instances per object type to allow direct comparison to the other method.  The result of our method on a scene with a larger number instances is shown in Figure~\ref{fig:teaser} and more examples are shown in the supplementary material.

Experiments on scenes with objects placed in an organized way gave the results seen in Figure~\ref{fig:duckthree}.  Here, three ducks were placed side by side and our algorithm ended up with a model of two repetitive patterns representing two neighboring ducks. This was expected as the clustering stage selects the largest cluster to start building the initial model.
Note that the filter eliminating overlapping triangles was turned off to create this example.


\subsection{Quantitative Results}

We report object discovery performance on the generated dataset in Table~\ref{table:results}. Objects were counted as true positives if the discovered model was correct, and it was localized correctly.  We considered a detection correct when at least 90\% of the features were inside the annotated bounding box.  They were counted as a false negative if the object was not detected.  False positives occur when an incorrect model is found, or when a model was found in an incorrect place. Since \cite{cho2009feature} only discovers pairwise matches, we compared the performance for the subset of scenes containing only pairs of objects (two object instances per object type).  Our method has an F1-score of $0.966$ on this subset, while the comparison method only reaches $0.471$. Our method finds any number of instances and thus can be evaluated on the entire dataset by counting the number of instances (not pairs) found. For this evaluation our F1-score is $0.974$.

Our method gives very few false positives resulting in a high precision.  A false positive example is shown in Figure~\ref{fig:falsepos}.  There is a few false negatives in our method (\eg, the brushes in the scene in row 5, column 1 of Figure~\ref{fig:dataset}).  These false negatives are caused by the objects having relatively few keypoints, and many of the keypoints being matched incorrectly.  These incorrect matches result from the descriptor not being sufficiently invariant to large viewpoint changes.  The recall does not differ much for the scenarios with or without occlusion and clutter, but it is slightly lower for the scenario with multiple object types.

An interesting result of the comparison method is that they have the worst performance for the easiest scenario $S_{ez}$ with only one object type without clutter or occlusion (\eg, the scenes in rows 1, 2 and 3, column 2 of Figure~\ref{fig:dataset}).  In this scenario, they have a large number of false positives.  As there are far fewer keypoints, it is more likely that false matches accidentally form a cluster.  If there is more clutter, these false matches are more likely to be more random and are less likely to cause false positive clusters.

\begin{table}[t]
\centering
\begin{tabular}{c|cc|cccc|}
\cline{2-7}
& \multicolumn{2}{|c|}{Objects (our)} & \multicolumn{2}{c}{Pairs (our)} &  \multicolumn{2}{c|}{Pairs~\cite{cho2009feature}}\\
\cline{2-7}
& P & R & P & R & P & R\\
\hline
$S_{ez}$& 0.992& 0.971& 1.000& 1.000& 0.338& 0.329\\
$S_{d}$& 1.000& 0.965& 1.000& 0.965& 0.621& 0.506\\
$M_{ez}$& 1.000& 0.919& 1.000& 0.931& 0.607& 0.508\\
$M_{d}$& 0.989& 0.907& 1.000& 0.925& 0.629& 0.429\\
\hline
Total& 0.995& 0.938& 1.000& 0.949& 0.504& 0.441\\
\hline
F1 & \multicolumn{2}{|c|}{0.974} & \multicolumn{2}{|c}{0.966} & \multicolumn{2}{c|}{0.471}\\
\hline
\end{tabular}
\caption{Precision (P) and recall (R) for our method and~\cite{cho2009feature} on the different datasets.  $S/M$ indicates whether the dataset has a Single (S) or Multiple (M) object models.  The easy/difficult scenario is indicated with $_{ez}$ or $_{d}$.  We give results for P-R calculated on the found object instances (first column) and for P-R calculated on pairs of objects (second and third columns).}
\label{table:results}
\end{table}

%% file: discussion.tex
\section{Conclusion and Discussion}

We presented a novel method for 3D discovery, modeling, and localization of multiple instances of an object using a single RGB-D image. Following a sparse feature representation, we employ appearance similarity and geometric similarity to group features associated to the instances. Our grouping algorithm is efficient as it considers triplet matches and eliminates incorrect correspondences between triplets based on various geometric constraints. The 6-DOF poses calculated for each triplet match are clustered in order to find matching object instances. The initial model generated using the clustering results can then be used to detect remaining object instances in the scene. The proposed method provides descriptive and compact object models using only a single RGB-D image and is suitable for robotic manipulation tasks. Another application of our framework can be seen in~\cite{Abbeloos2017}, where we generate object proposals for a deep-learning-based classification method. Since our technique outputs regions with recurrent patterns, we further improve classification accuracy by considering joint probability of bounding boxes that refer to copies of the same object.

The goal of this study is to detect and discover objects from a single image that contains multiple object instances. Therefore it differs from other methods that aim to detect and track objects in a given sequence of frames such as object SLAM techniques~\cite{Cadena2016}. Based on a sparse feature representation, the problem can be formulated as a challenging search problem, where we search for subsets of features that resemble each other in terms of appearance and geometry. More specifically, all pairwise feature matches need to be considered in this problem, as opposed to investigating feature matches only between frames in an object tracking scenario given a sequence of frames. Moreover, the instances can occur in various viewpoints in a single image making the search problem more challenging, while a smooth motion is usually observed in a video-based tracking scenario.

Our algorithm finds the largest recurrent pattern in the scene. This can be an important limitation especially in two cases (i) when the objects are placed in an organized way as in Figure~\ref{fig:duckthree}, and (ii) when there are pairs of instances that have the same relative pose. The limitation in the first case exists due to the fact that we avoided  any assumptions about the placement of the objects as opposed to existing work on finding periodic patterns in a scene. This makes our algorithm applicable to more general scenarios, i.e., occlusion, random pose, etc. Once a cluster is detected it is always possible to search for periodic patterns in order to handle cases with organized objects. We can also solve this problem by recursively calling the algorithm on the set of points from each cluster. On the other hand, if an object has repeated patterns on its surface, then following such a recursive splitting strategy would result in models smaller than the object. Consequently, such scenarios might be solved with more prior knowledge such as the size of the object or the number of object instances in the scene. The limitation in the second case can be solved by some assumptions about the placement of the objects. For example, a smoothness in depth can be expected among the features of an instance. Again, we avoided using any depth-based segmentation as they are shown to perform poorly on various cases.

Keypoint detection and descriptor matching lie at the core of the proposed  technique as a way of measuring appearance similarity. Missing features in some instances due to nonrobust keypoint detection and poor matching in large viewpoint changes were among the problems we faced because of limitations of conventional feature detectors and descriptors. As a result, the performance degraded especially when there are too many false matches that need to be handled as observed in the case of cluttered scene or multiple object types in a single image. Therefore, improving descriptors and descriptor matching is an important extension of this work.